# Extractive Text Summarization using Neural Networks

Aakash Sinha
Department of Computer Science
and Engineering
Indian Institute of Technology Delhi
New Delhi - 110016, India
cs1150202@iitd.ac.in

Abhishek Yadav
Department of Computer Science
and Engineering
Indian Institute of Technology Delhi
New Delhi - 110016, India
cs1150204@iitd.ac.in

Akshay Gahlot
Department of Computer Science
and Engineering
Indian Institute of Technology Delhi
New Delhi - 110016, India
cs1150209@iitd.ac.in

*Abstract*—Text Summarization has been an extensively studied problem. Traditional approaches to text summarization rely heavily on feature engineering. In contrast to this, we propose a fully data-driven approach using feedforward neural networks for single document summarization. We train and evaluate the model on standard DUC 2002 dataset which shows results comparable to the state of the art models. The proposed model is scalable and is able to produce the summary of arbitrarily sized documents by breaking the original document into fixed sized parts and then feeding it recursively to the network.

*Keywords—neural networks; recursive; extractive; summarization*

## I. INTRODUCTION

Text Summarization is a well-known task in natural language understanding. Summarization, in general, refers to the task of presenting information in a concise manner focusing on the most important parts of the data whilst preserving the meaning. The main idea of summarization is to find a subset of data which contains the "information" of the entire set. In today's world, data generation and consumption are exploding at an exponential rate. Due to this, text summarization has become the necessity of many applications such as search engine, business analysis, market review etc. Automatic Document summarization involves producing a summary of the given text document without any human help. This is broadly divided into two classes – Extractive Summarization and Abstractive Summarization. Extractive summarization picks up sentences directly from the document based on a scoring function to form a coherent summary. On the other hand, abstractive summarization tries to produce a bottom-up summary, parts of which may not appear as part of the original document. Such a summary might include verbal innovations although in most cases vocabulary of the summary is same as that of the original document. In general, building abstract summaries is a difficult task and involves complex language modeling. Text Summarization finds its applications in various NLP related tasks such as Question Answering, Text Classification, and other related fields. Generation of summaries is integrated into these systems as an intermediate stage which helps to reduce the length of the document. This, in turn, leads to faster access for information searching. News summarization and headline generation is another important application. Most of the search engines use machine-generated headlines for displaying news articles in feeds.

In this paper, we focus on extractive summarization. It focuses on extracting objects directly from the entire collection without modifying the objects themselves. Extractive summarizers take sentences as input and produce a probability vector as output. The entries of this vector represent the probability of the sentence being included in the summary. To produce the final summary best sentences are chosen according to the required summary length.

Various models based on graphs, linguistic scoring and machine learning have been proposed for this task till date. Most of these approaches model this problem as a classification problem which outputs whether to include the sentence in the summary or not. This is achieved using a standard Naive Bayes classifier or with Support Vector Machines [7,8,20]. Supervised learning based models rely on human-engineered features such as word position, sentence position, word frequency and many more. Based on these features each sentence is assigned a score. Various scoring functions including TF-IDF, centroid based metrics etc. [21,22]. have been used to date. Sentences are then ranked according to their importance and similarity using a ranking algorithm. The similarity between sentences can be calculated using cosine similarity. This is done to prevent the occurrence of repetitive information.

Feature engineering-based models have proved to be much more successful for domain or genre specific summarization (such as for medical reports or specific news articles), where classifiers can be trained to identify specific types of information. These techniques give poor results for general text summarization [8,23,24]. In this work, we propose a fully data-driven approach using neural networks which gives reliable results irrespective of the document type. This does not require predecided features for classifying the sentences. The proposed model is capable of producing summaries corresponding to documents of varying lengths. We have used a recursive approach to produce summaries of variable length documents. We trained the model using DUC datasets. We evaluated the proposed model using ROUGE automatic evaluator on DUC 2002 dataset and compare the ROUGE1 and ROUGE2 (two variants of ROUGE) scores with existing models. Experimental results show that the proposed model achieves performance comparable to state-of-the-art systems without any access to linguistic information.

Rest of the paper is presented as follows. In Section 2, we formulate the problem. Section 3 conceptualizes the proposed model and describes the neural network in detail. We have presented some information on datasets used and experimental details in Section 4. Comparison with various existing models has also been provided. The results of our experiments are shown in Section 5 and the paper is concluded in Section 6.

## II. PROBLEM FORMULATION

In this section, we describe the summarization task in a formal manner. Given a document X with a sequence of sentences {a1, a2, …, ax-1, ax}, we want to generate a summary at the sentence level. Extractive methods yield naturally grammatical summaries and require relatively little linguistic analysis. We create an extractive summary of the document by selecting a set of sentences {a1', a2',…, ay-1', ay'} from the document such that y<x i.e. the number of sentences in the summary is less than that in the original document. We will assume that the output length is fixed and the summarizer knows the length of the summary before generation. The selection process involves scoring each sentence in document X and predicting a label $w_L \in \{0,1\}$ which indicates whether the sentence should be included in the summary. Since we use a supervised learning technique, the objective is to maximize the likelihood of the sentence labels $\{w_{L1}, \ldots, w_{Lx}\}$ given the input document and model parameters ө.

$$\log p(w_L | X;ө) = \sum \log p(w_{Li} | X;ө) \quad (1)$$

For this purpose, a scoring function is used which assigns a value to each sentence denoting the probability with which it will get picked up in the summary. Because the summary length is fixed and known, top k (according to the summary length) sentences are chosen to be included in the summary. Thus we obtain an optimal 'k' sentence subset of the document which represents our summary. Quality of the summary depends upon the choice of these sentences.

## III. PROPOSED MODEL

The proposed model is based on a neural network which consists of one input layer, one hidden layer, and one output layer. The document is fed to the input layer, computations are carried in the hidden layer and an output is generated at the final layer. In this section, we talk about input vector generation, processing taking place in the network and summary generation from the output of the neural network.

Sentences of the document were to be fed as input to the network. Since the input to neural networks has to be numbers, the sentences have to be converted and represented in some numerical form. For this purpose, word2vec model was used. This model provides vector representation for words of the English language. A language model is trained on large datasets and each of the words in the vocabulary is assigned a vector of some fixed dimension based on the context in which it appears. Note that this dimension is fixed for each word. The model basically tries to predict the next word from given context words. These vectors have some important properties (for example closely related words have similar representations) which are more representative of the language. For more details on how these vectors are generated, the reader is advised to refer to word2vec by [14].

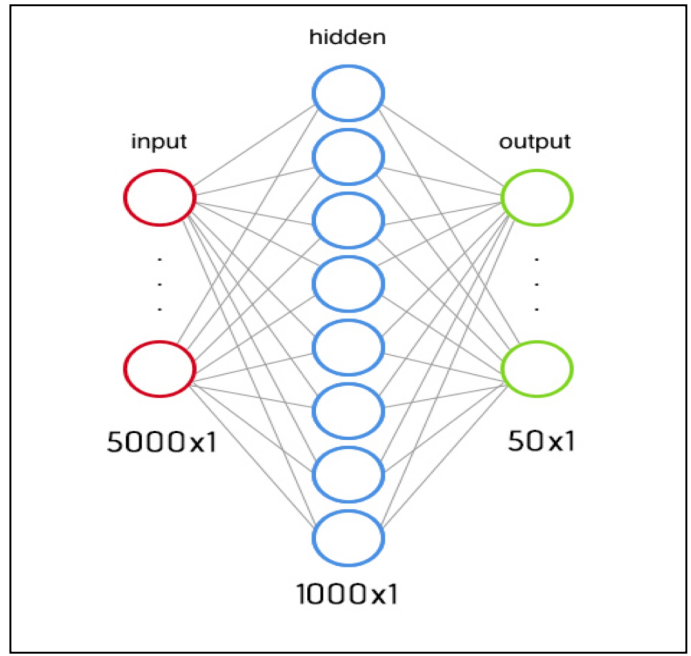

Fig. 1. Proposed Neural Network with 'page_len' = 50.

After obtaining the word vectors, vector representation for sentences had to be created. This representation should be such, that it is able to reflect the sentence in the best possible manner. One of the most intuitive approaches is that of averaging the word vectors. This doesn't turn out to be very useful and leads to poor results because of lack of consideration of order and relationship among the words. For generating a meaningful representation, some kind of contextual relation among the words has to be taken care of. For this, an approach based on n-grams was used. In this model, we used the Fasttext library [19] provided by Facebook to convert our sentences to vectors. The model takes input as sentences of the English language, vector representation of words and converts the sentences to fixed dimension vectors (100 in our case).

The size of the input layer is fixed and cannot be varied for different documents. Since each of the sentences has already been converted to fixed 100-dimensional vectors, we need not worry about variation in length of sentences. But one problem that still remains is that of variation in length of documents. Every document has different length in terms of the number of sentences and a summarizer should work well for all sizes. Because of this, various approaches using Recurrent Neural Networks and End to End learning have been proposed. Although they have been proven to work well, a lot of computation is needed for such models and they are fairly difficult to implement. Instead, we propose a simpler approach of summarizing the text recursively and show that the proposed model has a performance comparable to these complex systems.

Let the number of sentences in the document be 'doc_len'. Now we divide the document into segments, each having a fixed number of sentences. Each such segment is called a 'page' and let this fixed number be 'page_len'. In this way we obtain 'num_pg' pages, where 'num_pg' equals to ceil(doc_len/page_len). Thus, for each run of the network sentences of a page are converted into their corresponding vectors (each having 100 entries). All such vectors are concatenated in order to form a page_len*100-dimension vector which is fed to the input layer of the network. For pages with the number of sentences less than 'page_len', the input vector is padded with zeros. Note that 'page_len' is fixed for the model. Later on, we test the model for various values of this parameter and report the results.

A softmax activation function is applied to the output at the last layer. Each entry of the obtained vector denotes the weight associated with the corresponding sentence which represents the measure of belief of the sentence being included in the summary. Fig.1 shows the schematic representation of the model. As it is a supervised learning model, we already have the correct labels for the sentences of the document. Error/loss from the correct prediction is calculated using cross-entropy between the predicted output and the correct hot vector. This error is then fed back into the network for training. Thus the weights and bias matrices are adjusted in each iteration by back-propagating the error. An optimal value of the learning rate (the rate at which parameters are updated) is obtained through repeated experiments.

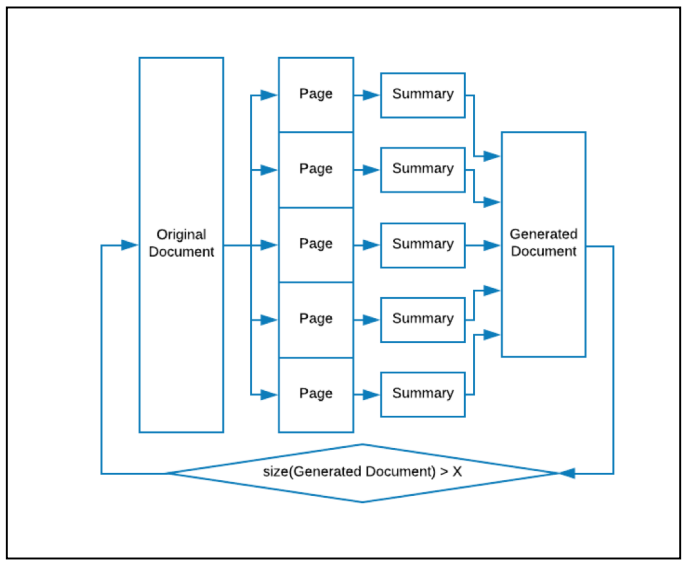

Fig. 2. Flow diagram of the proposed model.

For the generation of the summary of a given document, the entire text is broken into pages. The summary length in terms of the number of sentences is fixed and known before summary generation. Let this number be 'X'. Now each of the pages is fed to the network as an input. The network outputs a probability vector from which top 'X' sentences are chosen. Thus, a summary of length 'X' is generated for each of them. All such summary segments are concatenated in order to produce another document. This is then recursively fed to the summarizer till the number of sentences in the document reduces to 'X' (see Fig. 2). Thus, using this recursive approach we are able to generate a summary corresponding to the original document which consists of the best 'X' sentences and is a very good representative of the entire text. Using the final output vector, corresponding sentences are picked up from the document and concatenated in order to produce the final summary. Since this is a single document summarization model, we have assumed that 'X' will not exceed 'page_len'. Discussion on the value of 'page_len' is available further on in this paper.

IV. EXPERIMENTAL SETUP

In this section, we will explain how we measured the performance of the proposed network and, how we set the network parameters for optimal performance. We will also briefly discuss the dataset used for training and evaluation, and existing state of the art systems used for comparison.

A. Dataset

We trained the proposed model on two datasets. The first is the DUC 2002 datasets. The dataset in raw form consisted of XML pages which had to be pre-processed. The preprocessing involved converting the dataset into text documents. The dataset consisted of 567 document summary pairs divided into 59 clusters. Each document had two summaries, a 200-word summary and a 400-word summary (both extractive). We used the 200-word summaries for all purposes. For training, we used 75% documents and the rest were used for evaluating the model performance. We also evaluated the proposed model on a 35 document dataset used by [15]. This dataset also consisted of DUC documents and extractive summaries. Out of these 35 documents, we used 20 to train the network and the rest for evaluation purposes. We based the evaluation of the proposed model for the above training sets on two variants of ROUGE evaluator, namely ROUGE-1 and ROUGE-2. Further details of the evaluation and comparison with existing models can be found in the upcoming sections.

B. Implementation Details

We implemented the proposed model using TensorFlow library (tensorflow.org) which uses data flow graphs. TensorFlow allowed us to make the most of our available hardware. It is a flexible and portable library with Auto-Differentiation capabilities. Error/loss from the correct prediction was calculated using cross-entropy function. We tested various values of learning rate and hidden layer size to get the best combination of performance and computation time. This allowed us to get saturating accuracy after approximately 20 epochs. We fixed the number of sentences to be the 'page_len' parameter which denotes the number of sentences to be fed to the network at a time. Apart from the standard implementation mentioned above, we needed to set an appropriate value of the above parameter as well. To do this, we tested the performance of the proposed network using various values of the 'page_len' parameter on the entire DUC-2002 dataset. The performance comparison is shown using tables below.

TABLE I. PERFORMANCE OF THE PROPOSED NETWORK (ROUGE-1) ON DUC-2002 DATASET UPON VARYING THE 'PAGE_LEN' PARAMETER VALUE.

| page_len | Rouge-1 Score |
|---|---|
| 10 | 0.520 |
| 20 | 0.526 |
| 40 | 0.551 |
| 50 | 0.525 |
| 100 | 0.525 |
| 200 | 0.523 |

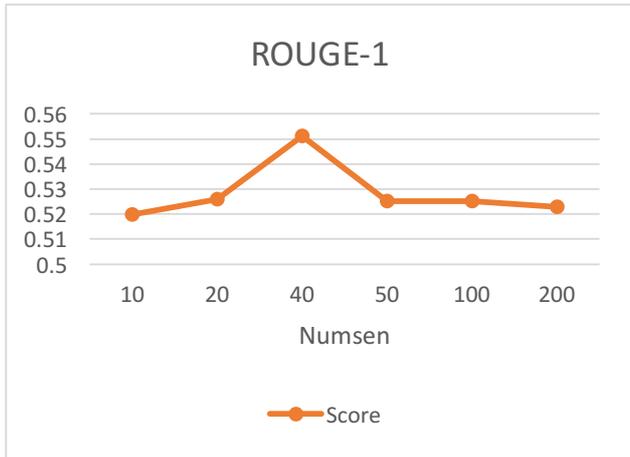

Fig. 3. Performance of network (ROUGE-1) on DUC-2002 dataset vs 'page_len' parameter.

TABLE II. PERFORMANCE OF THE PROPOSED NETWORK (ROUGE-2) ON DUC-2002 DATASET UPON VARYING THE 'PAGE_LEN' PARAMETER VALUE.

| page_len | Rouge-2 Score |
|---|---|
| 10 | 0.185 |
| 20 | 0.161 |
| 40 | 0.226 |
| 50 | 0.180 |
| 100 | 0.189 |
| 200 | 0.187 |

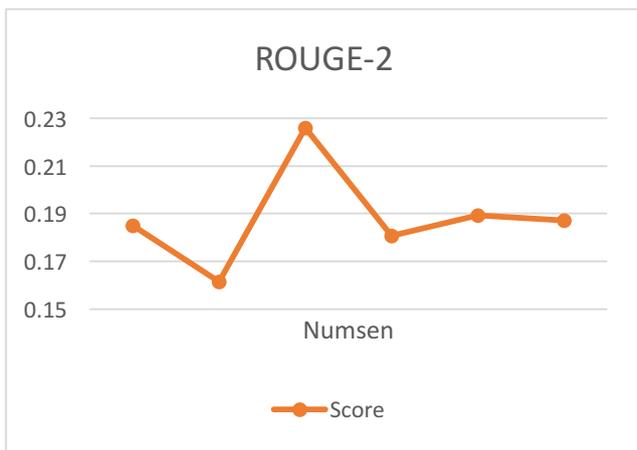

Fig. 4. Performance of network (ROUGE-2) on DUC-2002 dataset vs 'page_len' parameter.

The results show that the proposed network achieves maximum performance (considering both ROUGE-1 and ROUGE-2 results) when 'page_len' value is set to 40. That is, taking the document as input 40 sentences at a time gives best results. For more details of the ROUGE evaluator, refer to the upcoming sections.

*C. Performance Evaluation*

We compared the proposed model to various previously published models which are known to show good performance on the DUC-2002 dataset. The first one is a model is ILP by [16]. This model operates over a phrase-based representation of the source document which is obtained by merging information from PCFG parse trees and dependency graphs. Using an integer linear programming formulation, this model learns to select and combine phrases subject to length, coverage and grammar constraints. Another model is NN-SE by [1]. Their approach is based on Recurrent Neural Networks and shows very promising results. The next comparison system is TGRAPH [17]. This approach is based on a weighted graphical representation of documents obtained by topic modeling. Another system is URANK [18]. This model proposes a novel unified approach to simultaneous single-document and multi-document summarizations. The mutual influences between the two tasks are incorporated into a graph model and the ranking scores of a sentence for the two tasks are obtained in a unified ranking process. The last two comparison systems (namely TGRAPH and URANK) produce typical extractive summaries and are considered state-of-the-art. Finally, we also compared the proposed model with a system GENE proposed by [15]. This approach presents an extraction based single document text summarization technique using Genetic Algorithms.

V. RESULTS

In this section, we have shown how the proposed model faired against existing systems which are known to show good performance.

We used ROUGE for all evaluation purposes. ROUGE stands for Recall-Oriented Understudy for Gisting Evaluation. It is a measure which determines the quality of a summary automatically, by comparing it to human (ideal) generated summaries. Scores are allotted by counting the number of overlapping units between the computer-generated and the ideal summaries. The two variants of ROUGE used by us are ROUGE-1 and ROUGE-2. We have compared the proposed model with others on the basis of these two variants.

TABLE III. ROUGE SCORE (%) COMPARISON ON DUC-2002 DATASET (HIGHER IS BETTER)

| Model | Rouge-1 Score | Rouge-2 Score |
|---|---|---|
| ILP | 45.4 | 21.3 |
| TGRAPH | 48.1 | 24.3 |
| URANK | 48.5 | 21.5 |
| **Proposed Model** | 55.1 | 22.6 |
| NN-SE | 47.4 | 23.0 |

Table III, shows how the proposed model faired when compared to the models mentioned before. These models are known to show good performance on the DUC-2002 dataset but are based on complex approaches. They use Recurrent Neural Networks [1], sophisticated constraint optimization (ILP), sentence ranking mechanisms (URANK), etc. These approaches are hard to implement and require a lot of computation. On the other hand, our data-driven approach which uses a simple feedforward neural network is both implementationally and computationally light and obtains performance on par with state-of-the-art systems (evident from the table above).

TABLE IV. PRECISION COMPARISON BETWEEN THE PROPOSED MODEL AND GENE MODEL.

| Document Number | GENE (Precision) | Proposed Model |
|---|---|---|
| 1. | 0.5556 | 0.4445 |
| 2. | 0.6667 | 0.5556 |
| 3. | 0.6875 | *0.8750* |
| 4. | 0.7778 | 0.6667 |
| 5. | 0.6154 | *0.7692* |
| 6. | 0.6429 | *0.7143* |
| 7. | 0.7143 | 0.4300 |
| 8. | 0.6250 | *0.6250* |
| 9. | 0.6818 | *0.8182* |
| 10. | 0.6316 | *0.7900* |
| 11. | 0.7500 | *0.7500* |
| 12. | 0.8000 | *0.8000* |
| 13. | 0.7778 | *1.0000* |
| 14. | 0.7000 | 0.4000 |
| 15. | 0.7778 | *0.7778* |

Table IV shows the comparison of the proposed model's performance with the performance of a genetic algorithm based system GENE [15] on a 35 document dataset also used by [15]. The performance measure is a custom precision function used by [15] to demonstrate their systems performance. The table shows the performance of the proposed models using the same precision function. Results show that the proposed model easily outperforms this complex genetic algorithm approach as well.

VI. CONCLUSION AND FUTURE WORK

In this work, we presented a fully data-driven approach for automatic text summarization. We proposed and evaluated the model on standard datasets which show results comparable to the state of the art models without access to any linguistic information. We demonstrated that a straightforward and a relatively simpler approach (in terms of implementation and memory complexity) can produce results equivalent to complex deep networks/sequence-based models.

We have assumed that summary length to be generated should be less than 'page_len'. So we will try to improve upon this aspect in the future.